\documentclass[11pt,onecolumn] {IEEEtran} 
\ifCLASSINFOpdf
   \usepackage[pdftex]{graphicx}
\else
\fi
\usepackage[cmex10]{amsmath}
\usepackage{amssymb}
\usepackage{notacion2}

\ifx\pdfoutput\undefined
\usepackage{graphicx}
\else
\usepackage[pdftex]{graphicx}
\usepackage{epstopdf}
\fi
\usepackage[svgnames, x11names]{xcolor}

\usepackage{environ}

\NewEnviron{equationw}{
\begin{equation}\begin{split}
\BODY
\end{split}\end{equation}
}

  \usepackage[caption=false,font=normalsize,labelfont=sf,textfont=sf]{subfig}

\begin{document}
%
\title{Proper Complex Gaussian Processes for Regression}
%
%
%

\author{Rafael~Boloix-Tortosa*,
        F.~Javier~Pay\'an-Somet,
        Eva Arias-de-Reyna,
        and Juan~Jos\'e~Murillo-Fuentes
\thanks{R. Boloix-Tortosa, F.~Javier~Pay\'an-Somet, Eva Arias-de-Reyna,
        and Juan~Jos\'e~Murillo-Fuentes are with the Department of Signal Theory and Communications, University of Seville, Spain. e-mail: rboloix@us.es. Phone: +34 954488132.}
}

\hyphenation{op-tical net-works semi-conduc-tor hy-per-pa-ra-me-ters ge-ne-ra-tion ge-ne-ra-ted}

\maketitle

\begin{abstract}
Complex-valued signals are used in the modeling of many systems in engineering and science, hence being of fundamental interest. Often, random complex-valued signals are considered to be proper. A proper complex random variable or process is uncorrelated with its complex conjugate. This assumption is a good model of the underlying physics in many problems, and simplifies the computations. While linear processing and neural networks have been widely studied for these signals, the development of complex-valued nonlinear kernel approaches remains an open problem. In this paper we propose Gaussian processes for regression as a framework to develop 1) a solution for proper complex-valued kernel regression and 2) the design of the reproducing kernel for complex-valued inputs, using the convolutional approach for cross-covariances. In this design we pay attention to preserve, in the complex domain, the measure of similarity between near inputs. The hyperparameters of the kernel are learned maximizing the marginal likelihood using Wirtinger derivatives. Besides, the approach is connected to the multiple output learning scenario. In the experiments included, we first solve a proper complex Gaussian process where the cross-covariance does not cancel, a challenging scenario when dealing with proper complex signals. Then we successfully use these novel results to solve some problems previously proposed in the literature as benchmarks, reporting a remarkable improvement in the estimation error.

%
%
%

 \end{abstract}

\begin{IEEEkeywords}
Gaussian processes, regression, proper complex processes, kernel methods.
\end{IEEEkeywords}

%
\IEEEpeerreviewmaketitle

\setlength{\arraycolsep}{0.1em} 

\section{Introduction}
%
%
%
%

\IEEEPARstart{C}{omplex-valued} signals model a vast range of nowadays systems in science and engineering such as telecommunications, optics, electromagnetics, and acoustics among others. The main advantage of using complex-valued signals is the availability of processing the real and imaginary parts as a single signal. For these reasons, complex-valued signal processing is of fundamental interest.

Signal processing for complex-valued signals has been widely studied in the linear case, see \cite{Schreier06} and references therein. A proper complex random signal is uncorrelated with its complex conjugate \cite{Neeser93}. Properness is useful because it simplifies computations and in many cases models the underlying physics of the problem at hand. Most of the linear solutions in the real case admit a proper complex signals version with minor changes \cite{Tulay11}.  For the processing of improper signals further development of the linear solutions are needed \cite{Tulay11,Schreier06}.

The nonlinear processing of complex-valued signals has been addressed from the point of view of neural networks, \cite{hirose13} and, recently, using reproducing kernel Hilbert spaces (RKHS) \cite{Scholkopf02}. Some complex kernels have been lately proposed for classification  \cite{Steinwart06}, regression \cite{OgunfunmiP11,Bouboulis12,Tobar12} and mainly for kernel principal component analysis \cite{Papaioannou14}. 
Regarding regression, in \cite{Bouboulis11} the authors propose a complex-valued kernel based in the results in \cite{Steinwart06} and face the derivative of cost functions by using Wirtinger's derivatives. Same kernel is adopted in \cite{OgunfunmiP11}. As discussed later in this paper the resulting approach involves properness. Besides, the kernel used is neither stationary nor isotropic, and it may suffer from numerical problems. In \cite{Tobar12} the authors review the kernel design to improve the previous solution with a kernel they denote as independent. The resulting kernel yields also proper complex-valued outputs. The kernel is stationary, but again it is not isotropic in the complex-valued input space, as the real and imaginary parts of the input are split and fed to different real valued kernels. Hence, these previous works do not report results for isotropic and stationary kernels that may better model the underlying physics of some systems. Also, the structure of the kernel remains more rigid than needed. These drawbacks make these solutions not powerful enough to learn a wide range of systems. 
The previous approaches have been developed in the framework of kernel least squares and for adaptive filtering, although the extension to batch processing is immediate. We bring here Gaussian processes for regression (GPR) as a useful tool to better study this problem of kernel complex-valued nonlinear regression \cite{Boloix14}. 
%

GPRs are well known techniques in machine learning \cite{Rasmussen06,PerezCruz13gp}. They have been successfully applied to regression, classification and dimensionality reduction. GPs can be interpreted as a family of kernel methods with the additional advantage of providing a full conditional statistical description for the predicted variable, where hyperparameters can be learned by maximizing the marginal likelihood, avoiding cross-validation. 
Properties and features of random complex-valued Gaussian processes have been widely studied \cite{Miller69}, facilitating the extension of GPR for the complex-valued case proposed in this paper.  The benefits are twofold. On the one hand we have a novel tool for nonlinear regression providing probabilistic outputs. Second, since this Bayesian approach is connected to the regularization point of view \cite{Alvarez12}, we prove the structure of the reproducing kernel and study the conditions to be valid for proper complex outputs. 
The solution of GPR in the general complex-valued case is quite involved, since the pseudo-covariance is to be taken into account. In the proper scenario the pseudo-covariance cancels, and the formulation of Gaussian processes for real-valued signals can be adapted \cite{Neeser93}. We describe this new solution, providing the mean and the covariance of the complex-valued outputs. 
We particularly focus on the kernel, to improve previous solutions. We derive its structure, being complex valued, where the real part of the kernel is given by the covariance of the real part of the outputs plus the covariance of the imaginary part of the outputs, while the imaginary part of the kernel describes the cross-covariance between real and imaginary parts of the outputs. We prove that the real and imaginary parts of the kernel can be designed with different features. But we conclude that the imaginary part, in addition to be skew-symmetric, must be constructed to ensure the whole covariance to be semi-definite positive, i.e. a reproducing kernel or covariance matrix \cite{Taubock12}. 
We also pay attention to the modeling of physical systems, and propose a kernel for the real part based on the exponential kernel. At this point, when measuring similarity between inputs we resort to the absolute value of the complex difference between inputs. As a result, the kernel is isotropic and stationary. Other real kernels may be adapted using this metric, to better model the physics of the problem at hand. 
The construction of the imaginary part is challenging. Conditions on this matrix to ensure properness and to be a valid kernel quite limit the degree of freedom in the kernel design, and the systems we may model. We resort to the convolution approach \cite{Boyle05,Calder07} to ensure the whole matrix to be a reproducing kernel. We propose a model for the cross-covariance that explains a positive and negative correlation of the real part of the output with the imaginary one for a positive and negative delay, canceling at origin.   
Every system modeled with complex-valued signals can be rewritten with real-valued ones by splitting the output signal into a composite two-dimensional vector, with real and imaginary parts. For the sake of comparison, the multiple output learning (MOL) \cite{Boyle05,Alvarez12} is proposed to solve this composite kernel complex-valued regression problem and adapt it to the proper complex case. In this scenario we lose the intuition about complex values, and we cannot deal with inputs and outputs easily pairing real and imaginary parts to measure similarity between them. Also, we may lose some of the computational benefits of working with complex numbers,  
such as in the computation of the matrix inversion,  \cite{Ehrlich70}, \cite{Golub96}. 
However, we prove that both approaches yield equivalent results, gaining some further intuition on the kernel properties and structure.   

One of the advantages of GPR is the availability of estimating hyperparameters by maximizing the marginal likelihood. Here we face the derivation of a function of complex-valued parameters and variables. We cope with this problem by using Wirtinger's derivatives. 
To illustrate the good performance of this approach, we include a first experiment where we learn a Gaussian process with unknown hyperparameters, where we assume the cross-covariance does not cancel. Then we compare to the algorithms in \cite{Bouboulis12,Tobar12} using the same scenario than in the experiments described in \cite{Bouboulis12} as benchmark. Due to the improvements derived from the proposals in this paper, we achieve remarkable good results. At this point we conjecture that whenever we do not know the cause or model of the cross-covariance or this cross-covariance is negligible, assuming a real-valued reproducing kernel is a good option.




The rest of the paper is organized as follows. Next section includes the derivation of GPR for proper complex-valued signals. In \SEC{Cov} we exploit this formulation to study the structure and construction of the kernel, compare it to the kernel in the MOL approach, and provide some examples for the kernel. 
We develop in \SEC{ML} the optimization procedure to set the kernel hyperparameters applying Wirtinger's calculus and patterned complex-valued matrix derivatives \cite{Hjorungnes08}. The good performance of this approach is illustrated in \SEC{Exp}, learning a unknown process and comparing to previous approaches in the channel equalization problem. We end with conclusions and an appendix.

The notations used in the paper are as follows. If $\vect{A}$ is a matrix, $\entry{\vect{A}}{l}{q}$ is its $(l,q)$ entry. For a vector, $\vect{a}$, $a_l$ denotes its $l$-th entry.
$\vect{A}\trs$ represents the transpose of $\vect{A}$ and $\vect{A}\her$ its Hermitian. 
$\trace{\vect{A}}$ is the trace of $\vect{A}$, and $\det{\vect{A}}$ its determinant. 
To denote the $i$-th sample we use $(i)$, both for a vector, $\vect{a}(i)$, and a scalar, $a(i)$. 
${\vect{a}}_n$ is a vector of $n$ entries while $\vect{A}_n$ is a matrix with $n$ columns. The real and imaginary parts are denoted by subindex $\textrm{r}$ and $\textrm{j}$, respectively, i.e. $\vect{a}\rr=\Re(\vect{a})$ and $\vect{a}\jj=\Im(\vect{a})$, and throughout the text $\textrm{j}=\sqrt{-1}$. To denote the complex Gaussian distribution with mean vector $\boldsymbol{\mu}$, covariance matrix $\K$ and pseudo-covariance matrix $\matr{\tilde{K}}$ we use $\calg{N}\left(\boldsymbol{\mu},\K,\matr{\tilde{K}}\right)$.
$\E[{\cdot}]$ denotes statistical expectation. 


\section{Proper Complex Gaussian Process Regression}\LABSEC{GPRC}
GP for regression can be presented as a nonlinear regressor that expresses the input-output relation through function $f(\x)$, known as latent function, that follows a GP and underlies the regression problem
\begin{equation}\LABEQ{regression}
{\y}=\f(\x)+\epsilon,
\end{equation}
where the error, $\epsilon$, in the estimation of the output, $\y$, is modeled as additive zero-mean Gaussian noise.
GPs are the natural nonlinear Bayesian extension to the linear minimum mean-squared error (LMMSE) and Wiener filtering \cite{MurilloGP09,PerezCruz13gp}, thus, GPs provide the correct approach to solve an MMSE filter nonlinearly. 

For any input set $\{ (\x(i)) | i = 1, . . . , \n\}$, where $\x({i}) \in {\CN}^d$ are complex-valued column vectors of dimension $\d$, the latent function in \EQ{regression} can be designed and evaluated to provide a multidimensional
Gaussian complex-valued random vector $\fv_\n=\left[{f}(\x({1})), . . . , {f}(\x(\n))\right]^{\top}$, where $f(\x(i))\in \CN$. In this paper we focus on both complex-valued input and output. The simpler real-valued input and complex-valued output case can be easily solved from the results herein. 
The immediate approach when dealing with complex-valued signals is to consider the real and imaginary components as separate real signals \cite{PerezCruz08b}, for both inputs and outputs. If we wish real and imaginary parts to be considered dependent, we may resort to multiple output Gaussian processes to relate them \cite{Boyle05,Alvarez12}. However, in these approaches we lose the complex value notion,  specially for the input.


A complex Gaussian vector is characterized not only by its mean vector $\boldsymbol{\mu}$ and covariance matrix $\K= \mathbb{E}\left[(\fv_\n-\boldsymbol{\mu})(\fv_\n-\boldsymbol{\mu})\her\right]$, but also by its complementary covariance or pseudo-covariance matrix $\matr{\tilde{K}}=\mathbb{E}\left[(\fv_\n-\boldsymbol{\mu})(\fv_\n-\boldsymbol{\mu})^\top\right]$, \cite{Schreier06}. We use the notation $\calg{N}\left(\boldsymbol{\mu},\K,\matr{\tilde{K}}\right)$ for a joint Gaussian probability distribution to emphasize this fact.
When the complementary covariance matrix is zero, a complex Gaussian random vector is regarded as \emph{proper} \cite{Neeser93,Ollila08}. 
Without loss of generality, we consider here a zero-mean process for the latent function \cite{Neeser93}. By imposing properness, $(\psc{\K})_{il}=\psc{\k}(\x(i),\x(l))=0$ for all $i,l$. The prior for this proper complex Gaussian process yields 
\begin{equationw}\label{eq:propergaussian}
p(\fv_\n|\X_\n)=\calg{N}\left(\vect{0},\K,\matr{0}\right)=\frac{1}{\pi^n \det{\K}}\exp{\left(-{\fv}_\n\her{\K}\inv{\fv}_\n\right)},
\end{equationw}
where $\X_\n=\left[\x({1}), . . . , \x(\n)\right]$, and $({\K})_{il}={\k}(\x(i),\x(l))$ for all  $i,l$ in the input set, being ${\k}(\x(i),\x(l))$ the covariance function.

In the learning process we condition the output of the GPR for some new observations 
given the training set $\tset=\{(\x(i),y({i})) | i = 1, . . . , \n\}=\{\X_\n,\yv_\n\}$, where the outputs  $\yv_\n=\left[{\y}({1}), . . . , {\y}(\n)\right]^{\top}$, $\y(i)\in \CN$, for a given set of observations $\X_\n$, are known. We can compute this posterior through the joint distribution of the to-be-estimated outputs and the training data $\yv_\n$. This distribution can be constructed by using \EQ{propergaussian} and the marginal likelihood, $p\left(\yv_\n| \X_\n \right)$, as follows. 
We assume that the additive noise $\epsilon$ in \EQ{regression} follows an i.i.d. proper (circular) complex Gaussian distribution with zero mean and variance $\sigma^2_\epsilon$.
The samples in the training set are i.i.d., hence the likelihood for the latent function at the training set, $p\left(\yv_\n|\fv_\n \right)$, is given by a factorized model
\begin{equation}
p\left(\yv_\n| \fv_\n \right)= \prod_{i=1}^n {p(y({i})|f(\x({i})))},
\end{equation}
where $p(y({i})|f(\x({i})))=\calg{N}(f(\x({i})),\sigma^2_\epsilon,0)$.
Therefore, the likelihood is a proper complex multidimensional Gaussian $p\left(\yv_\n| \fv_\n \right)= \calg{N}\left(\fv_\n,\sigma^2_\epsilon\id{\n},\matr{0}\right)$. This likelihood and the prior in \EQ{propergaussian} yield the marginal likelihood or evidence 
\begin{equation}\label{eq:marlikelihood}
p\left(\yv_\n| \X_\n \right)=\int{p\left(\yv_\n| \fv_\n \right)p(\fv_\n|\X_\n)d\fv_\n} =\calg{N}\left(\vect{0},\C,\matr{0}\right),
\end{equation}
where  $\C=\K+\sigma_\epsilon^2\id{\n}$.
Note that $\yv_\n$ is also proper Gaussian. Furthermore, $\yv_\n$ and $\fv_\n$ are cross-proper, as the complementary cross-covariance matrix $\mathbb{E}\left[\yv_\n\fv_\n^\top\right]=\matr{0}$.
Hence, $\yv_\n$ and $\fv_\n$ are jointly proper \cite{Schreier06}, i.e., the composite complex random vector $\left[\yv_\n^\top,\fv_\n^\top\right]^\top$ is Gaussian proper. If we now have a number $m$ of test input vectors ${\X\new }_{m}=\left[\x\new({1}), . . . , \x\new({m})\right]$, the joint distribution of the training outputs $\yv_\n$ and ${\fv\new}_{m}=\left[{f}(\x\new({1})), . . . , {f}(\x\new({m}))\right]^{\top}$ is as follows: 
\begin{equation}\label{eq:yf}
\left[ \begin{array}{c}
\yv_\n\\
{\fv\new}_{m}\\
\end{array}\right]\sim\calg{N}\left( \matr{0}, \left[ \begin{array}{ccc}
\C & {} &\K(\X_\n,{\X\new }_{m})\\
\K({\X\new }_{m},\X_\n) & {} & \K({\X\new }_{m},{\X\new }_{m})\\
\end{array}\right],\vect{0}\right),
\end{equation}
where the entry $(i,l)$ of matrix $\K(\X_\n,{\X\new }_{m})$ is $\k(\x({i}),\x\new({l}))$, and in a similar way are defined $\K({\X\new }_{m},\X_\n)$ and $\K({\X\new }_{m},{\X\new }_{m})$. From these results, the estimated probabilistic output is the conditional distribution of ${\fv\new}_{m}$ given $\yv_\n$:
\begin{equation} \label{eq:posteriorf}
p\left({\fv\new}_{m}| {\X\new }_{m},\X_\n,\yv_\n\right)=\calg{N}\left(\boldsymbol{\mu}_{{\fv\new}_{m}},\matr{\Sigma}_{{\fv\new}_{m}},\matr{0}\right),
\end{equation}
where 
\begin{align}
\boldsymbol{\mu}_{{\fv\new}_{m}}&=\K({\X\new }_{m},\X_\n)\C\inv\yv_\n, \LABEQ{mediaCPGP}\\
\matr{\Sigma}_{{\fv\new}_{m}}&=\K({\X\new }_{m},{\X\new }_{m})-\K({\X\new }_{m},\X_\n)\C\inv\K(\X_\n,{\X\new }_{m}).\label{eq:varCPGP}
\end{align}

The predictive distribution is also proper Gaussian. 
The predictive covariance is a quadratic form of the test and training inputs, showing that the predictive uncertainties grow with the magnitude of the covariance matrices involved, as one would expect for a linear model.  



%


\section{Complex Covariance Functions}\LABSEC{Cov}
The covariance function is a key tool in GPR: it encodes our assumptions about the function that we wish to learn and measures similarity between inputs. There are some well known examples of covariance functions used for real-valued applications, such as the squared exponential or Gaussian covariance function and the inhomogeneous polynomial kernel, among others. These kernels assume that training points that are near to a test point should be informative about the prediction at that point \cite{Rasmussen06}. The definition of a kernel for complex-valued GPR should provide the same measure of similarity between the inputs. 

In the proper complex-valued Gaussian processes framework we can derive the kernel following a more principled approach as follows.
Consider a zero-mean complex Gaussian vector $\fv_\n={\fv_\n}\rr+\j{\fv_\n}\jj$, with ${\fv_\n}\rr$ its real part and ${\fv_\n}\jj$ its imaginary part. The covariance matrix $\K= \mathbb{E}\left[\fv_\n\fv_\n\her\right]$ is \cite{Schreier06}: 
\begin{equation}\label{eq:compleskernel}
\K=\K\rrrr+\K\jjjj+\j\left(\K\jjrr-\K\rrjj\right),
\end{equation}
where $\K\rrrr$ and $\K\jjjj$ $\in\RN_+^{n\times n}$ are the covariance matrices of real and imaginary parts of $\fv_\n$, respectively, and $\K\rrjj= \mathbb{E}\left[{\fv_\n}\rr{\fv_\n}\jj^\top\right]=\K\jjrr\trs$ $\in\RN^{n\times n}$ is the cross-covariance matrix of real and imaginary parts. At this point it is most important to notice that $\K$ must be a valid covariance matrix for the Gaussian process. Real-valued covariance functions can now be used to write out the three real covariance matrices $\K\rrrr$, $\K\jjjj$ and $\K\rrjj$, but they are interrelated. Given two inputs, the complex covariance function of the process is composed by three real covariance functions, 
\begin{equation}\label{eq:complexCV}
\k(\x,\x')=k\rrrr(\x,\x')+k\jjjj(\x,\x')+\j\left(k\rrjj(\x',\x)-k\rrjj(\x,\x')\right).
\end{equation}

For the particular case of a proper complex Gaussian vector, $\K\rrrr=\K\jjjj$ and $\K\jjrr=\K\rrjj\trs=-\K\rrjj$ \cite{Schreier06}. Therefore, in the case of a proper complex Gaussian process, functions $k\rrrr(\x,\x')=k\jjjj(\x,\x')$, and $k\rrjj(\x,\x')$ must yield either a null or skew-symmetric cross-covariance matrix $\K\rrjj$. The covariance matrix yields
\begin{equation}\LABEQ{complesProperkernel}
\K=\K\rr +\j\K\jj,
\end{equation}
where $\K\rr =2\K\rrrr$ and $\K\jj=-2\K\rrjj$. Following the guidelines in \cite{Alvarez12} we may conclude that the proposed kernel in \EQ{complesProperkernel} is a valid reproducing kernel if it is a covariance matrix for the Bayesian Gaussian process framework. Hence, it must be a Hermitian positive semi-definite matrix \cite{Taubock12}. $\K\rrrr$ must be a symmetric and positive semi-definite matrix, since the marginals of the joint probability functions of the real and imaginary parts must be also covariance matrices as discussed later in this section. It follows that the condition $\vect{v}\her \K \vect{v}\geq0$ for any $\vect{v}\in\CN$ yields,
\begin{align}\LABEQ{kercon}
\vect{v}\her \K \vect{v}=\vect{v}\rr\trs \K\rr \vect{v}\rr+\vect{v}\jj\trs \K\rr \vect{v}\jj-2 \vect{v}\rr\trs \K\jj \vect{v}\jj \geq 0
\end{align}
where the first terms to the right of the equality are greater or equal to cero, since $\K\rr$ is positive semi-definite and we used $\vect{v}\rr\trs \K\jj \vect{v}\rr=$$\vect{v}\jj\trs \K\jj \vect{v}\jj=0$ since $\K\jj$ is skew-symmetric. 
In the following subsection we analyze this problem by decomposing it into the real and imaginary parts.


\subsection{Complex GPR as multiple output GPR}\LABSEC{MO}

%

The complex case may be tackled by mapping the complex value into a two dimensional vector with real and imaginary parts, as in \cite{PerezCruz08}. Then two GPs can be learned, one for the real part and another for the imaginary part of the output. Either independently or using a MOL or vector scheme \cite{Micchelli05,Boyle05,Alvarez12}. The definition of RKHS valued functions in MOL parallels the one in the scalar, where now the reproducing kernel is matrix valued. In our problem and for the training set, we have scalar reproducing kernels $\K\rrrr$, $\K\rrjj$, $\K\jjrr$ and $\K\jjjj$, that are block matrices in the vector kernel matrix
\begin{align} \label{eq:Cmo}
\K_{\Rext} = 
\left[\begin{array}{c c}
\K\rrrr(\X_\n,\X_\n) & \K\rrjj(\X_\n,\X_\n) \\
\K\jjrr(\X_\n,\X_\n) & \K\rrrr(\X_\n,\X_\n)
\end{array}\right].
\end{align} 

For one single test input, we write the mean of the multiple outputs as follows
\begin{align} \label{eq:mimo}
\left[\begin{array}{c}
\hat{\y}\rr \\ \hat{\y}\jj
\end{array}\right]
= 
\left[\begin{array}{c c}
\kv\rrrr({\x\new},\X_\n) & \kv\rrjj({\x\new},\X_\n) \\
\kv\jjrr({\x\new},\X_\n) & \kv\jjjj({\x\new},\X_\n) 
\end{array}\right]
\C_{\Rext}\inv
\left[\begin{array}{c}
{\yv}\rr \\ {\yv}\jj
\end{array}\right]
\end{align}
where $\C_{\Rext}=\K_{\Rext}+\sigma_\epsilon^2\id{2\n}$ is the covariance matrix in the multiple output or vector formulation.

If we assume the output to be proper the model above yields
\begin{align} \label{eq:mimopr}
\left[\begin{array}{c}
\hat{\y}\rr \\ \hat{\y}\jj
\end{array}\right]
= 
\left[\begin{array}{c c}
\kv\rrrr({\x\new},\X) & \kv\rrjj({\x\new},\X) \\
-\kv\rrjj({\x\new},\X) & \kv\rrrr({\x\new},\X) 
\end{array}\right]
\left[\begin{array}{c c}
\C\rrrr & \C\rrjj \\
-\C\rrjj & \C\rrrr
\end{array}\right]\inv
\left[\begin{array}{c}
{\yv}\rr \\ {\yv}\jj
\end{array}\right].
\end{align}
where, given that the vector covariance matrix must be symmetric, $-\C\rrjj=\C\rrjj^T$. This formulation yields the solution in \EQ{mediaCPGP} with the kernel as proposed in \EQ{complesProperkernel}. Similar conclusions can be drawn for the covariance of the output in \EQ{varCPGP}. This result for the multiple output case links the resulting kernel in the complex value domain to those defined in the MOL case. Since matrix $\C_{\Rext}$
is the covariance matrix of the multiple output, it must be Hermitian and positive semi-definite, i.e., equation \EQ{kercon} must hold true for any $\vect{v}$. Also, since the marginal for the real or the imaginary part of the output is also a Gaussian process, matrices $\K\rrrr$ and $\K\rrjj$ must also be symmetric and positive semi-definite. 

If, in addition,
real and imaginary parts are uncorrelated for any pair of inputs, $\C\rrjj=0$, and \EQ{mimopr} yields
\begin{align} \LABEQ{propind}
\left[\begin{array}{c}
\hat{\y}\rr \\ \hat{\y}\jj
\end{array}\right]
= 
\left[\begin{array}{c}
\kv\rrrr({\x\new},\X)\C\rrrr\inv{\yv}\rr \\ \kv\rrrr({\x\new},\X)\C\rrrr\inv{\yv}\jj
\end{array}\right].
\end{align}  
This result can be easily derived from \EQ{mediaCPGP} with $\K=\K\rr=2\K\rrrr$ in \EQ{complesProperkernel}. Here, \EQ{kercon} always holds true if $\K\rrrr$ is positive semi-definite.

The multiple output formulation is equivalent to the complex-valued one. However, the notion of complex signals is lost. Also, if we split the inputs into real and imaginary parts, the design of the covariance matrix becomes a harder task if we want to measure similarity between complex signals.  In contrast, the complex formulation for GPR presented above allows for a kernel design in a natural way, better managing the complex nature of the output and the inputs, as follows.

%


\subsection{Design of Covariance Matrices}

In \cite{Steinwart06,Bouboulis11} it is proposed a complex-valued Gaussian kernel as an extension of the real Gaussian kernel: 
\begin{align}\LABEQ{RK}
&\k_{\CN}(\x,\x')=\exp\left(-({\x-\x'^{*})}^\top(\x-\x'^{*})/\gamma\right)\nonumber\\
&=\exp\left(-(|\x\rr-\x\rr'|^2/\gamma\right)\exp\left(|\x\jj+\x\jj'|^2/\gamma\right)\cdot\left(\cos(2(\x\rr-\x\rr')\trs(\x\jj+\x\jj')/\gamma)-\j\sin(2(\x\rr-\x\rr')\trs(\x\jj+\x\jj')/\gamma)\right),
\end{align}
where $\x=\x\rr+\j\x\jj$, $\x'=\x'\rr+\j\x'\jj$.
In \cite{Tobar12}, the authors propose the so-called independent kernel to improve the previous one: 
\begin{equation}\LABEQ{indepCGK}
\k_{ind}(\x,\x')=\kappa_{\Rext}\left(\x\rr,\x'\rr\right)+\kappa_{\Rext}\left(\x\jj,\x'\jj\right)+\j\left(\kappa_{\Rext}\left(\x\rr,\x'\jj\right)-\kappa_{\Rext}\left(\x\jj,\x'\rr\right)\right),
\end{equation}
where $\kappa_{\Rext}$ is a real kernel of real inputs. 

In \EQ{RK} we have that the value of the kernel for $(\x,\x')$ is the complex conjugate of the kernel for $(\x',\x)$. This corresponds to assuming the output is proper with non-null skew-symmetric cross-covariance matrix. 
This kernel measures similarities between real parts while measures dissimilarity between imaginary ones and it is not stationary. It also has an oscillatory behavior. In addition, the exponent in the kernel may easily grow large and positive, causing numerical problems, as later discussed in the experiments. These characteristics are not very useful when modeling the underlying physics of many systems.

The kernel in \EQ{indepCGK} solves the measure of similarity between inputs by checking for the real and imaginary parts independently. At this point, we lose the intuition about the complex nature of inputs. The kernel assumes again that the output process to model is proper complex valued, where the imaginary part is non-null and skew-symmetric. One of the main drawbacks of this kernel is that it is not isotropic, due to the way real and imaginary parts of the inputs have been split in the kernel. For example, if a real exponential kernel is used in \EQ{indepCGK} as proposed in \cite{Tobar12}, $\kappa_{\Rext}=\alpha\exp(-|\x-\x'|/\beta)$ for some hyperparameters $\alpha$ and $\beta$, whenever two inputs are distant enough the kernel vanishes except for similar imaginary parts. For any $\x\jj=\x\jj'$, the covariance yields the maximum value for $\kappa_{\Rext}$, $\k_{ind}(\x,\x')=\alpha$. In the imaginary part of the kernel we have a similar behavior. 



The proposed kernel in \EQ{complexCV} much better adapts to the problem at hand. We may design the real and the imaginary parts of the kernel with different structure. And if the cross-covariance is null or negligible we may set the imaginary part of the kernel to zero. 
The conditions for these kernel functions to form a reproducing kernel are given in \EQ{kercon}. On the other hand, we propose as measurement of similarity or metric the inner product, $\x\her\x$, used in the complex value space as follows.

\subsubsection{Real kernels} 

Assume the cross-covariance cancels and we resort to \EQ{propind}. We do use a \emph{proper} complex Gaussian process to model the system. But we later condition to the training samples. The resulting output within the input range defined by the training inputs may behave, locally, quite differently from a proper Gaussian process with null cross-covariance. Note that in \EQ{propind} we have the same matrix multiplying the real and imaginary parts of the outputs. If these parts are dependent, the GPR will translate this dependence to the output. Furthermore, if the real part of the output has a different variance than the imaginary part, \EQ{propind} properly scales the estimated output, by using the variance of the training outputs. For these reasons, we conjecture that if we know little about the cross-covariance between real and imaginary parts or it is negligible, using \EQ{propind} is a good choice as long as the complex nature of the inputs is properly addressed.

As already discussed, we propose using the inner product of the inputs, $\x\her\x$, a simple metric in complex numbers, to cope with isotropy. The Gaussian kernel yields
\begin{equation} \LABEQ{expkernel}
\k_{G}(\x,\x')=\exp\left(-(\x-\x')\her(\x-\x')/\gamma\right),
\end{equation}
that it is stationary. Therefore, a suitable choice for 
$k\rrrr(\x,\x')=k\jjjj(\x,\x')$ 
can be $\k_{G}(\x,\x')$ in \EQ{expkernel} while $k\rrjj(\x,\x')$ could be set to zero if we assume that the cross-covariance between real and imaginary parts cancels. If any further information on the physics of the model is known, other kernels \cite{Rasmussen06} can be derived using this metric. 
%

\subsubsection{Complex-valued kernels}The design of the imaginary part of the kernel in \EQ{complexCV} is a challenging problem that remains open, as it is quite system dependent. Here we propose just one kernel, but the procedure followed could help in other designs. The design should met condition \EQ{kercon}. This condition is not constructive, i.e. it does not help designing the kernel. But it encodes intuitive facts such as the maximum absolute value of the cross-covariance being lower or equal to the maximum absolute value of the covariance. On the other hand, the kernel must be able to explain the dependencies between real and imaginary parts of the output, if known. And we are restricted to skew-symmetric matrices. 

We propose to model a system where the real and imaginary parts of the outputs are correlated for delayed points with delay $\mu \in \CN^\d$. Since the covariance matrix must be skew symmetric, the correlation will be positive (or negative) for a delay of $\mu$ and negative (or positive) for a delay of $-\mu$. We must ensure that the kernel corresponds to a covariance matrix. We bring here the convolution approach \cite{Boyle05,Calder07}. We model the output of the process as the filtering of two independent white Gaussian noises, $S\rr$ and $S\jj$, and compute the kernels from the filter responses. This way we met the condition of being a covariance matrix. The filters are designed to model the proposed system, fulfilling the imaginary part of the kernel being skew-symmetric. The convolution process is sketched in Fig. 1 where we use exponential responses for the filters,
\begin{align}\LABEQ{fil}
h\rr(\x)&=h\jj(\x)=v\rr\exp(-(\x)\her(\x)/\gamma)\nonumber\\
h\rrjj(\x)&=v\rrjj\exp(-(\x-\mu)\her(\x-\mu)/\gamma)\nonumber\\
h\jjrr(\x)&=-v\rrjj\exp(-(\x+\mu)\her(\x+\mu)/\gamma)
\end{align}
and $W$ is some zero-mean stationary Gaussian noise with variance $\sigma_W^2$. For these filters, it can be proved, see the Appendix, that up to a multiplying constant the kernel in \EQ{complexCV} yields,
\begin{align}\LABEQ{kerfil}
\k(\x, \x')&=(v\rr^{2}+v\rrjj^{2})\exp\left(-\frac{\mathbf{d}_\x\her\mathbf{d}_\x}{2\gamma}\right)\nonumber\\
&+\j v\rr v\rrjj\left(\exp\left(-\frac{(\mathbf{d}_\x-\mu)\her(\mathbf{d}_\x-\mu)}{2\gamma}\right)\exp\left(-\frac{(\mathbf{d}_\x+\mu)\her(\mathbf{d}_\x+\mu)}{2\gamma}\right)\right).
\end{align}
where $\mathbf{d}_\x=\x'-\x$. Note that the constants $v\rrjj$ $\in \RN$, $v\rr$ $\in \RN$ and $\mu$ $\in \CN^\d$ could be set according to the problem at hand or learned as hyperparameters. Also, note that kernel in \EQ{expkernel} is a particularization of \EQ{kerfil}, by setting $v\rrjj=0$.

\begin{figure}[tb!]
\begin{center}
\includegraphics[width=6cm, draft=false]{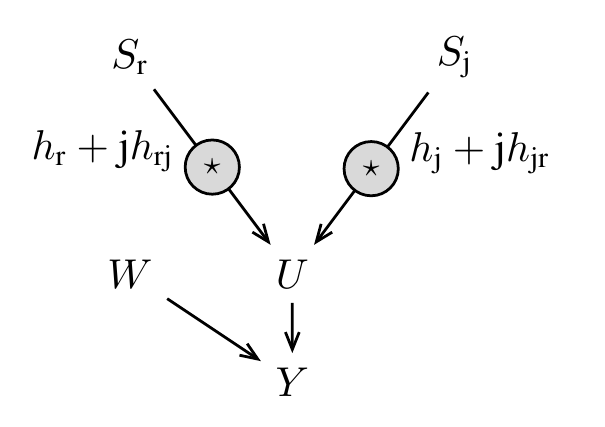}
\end{center}
\vspace*{-.6cm}
\caption{Convolution system model to design a kernel for proper Gaussian processes.}
\end{figure}

\section{Hyperparameters computation}\LABSEC{ML}


In order to turn complex GPR into a powerful practical tool we must address the model selection problem, in the sense of setting the hyperparameters. A major advantage of GPR is that the hyperparameters can be estimated by maximizing the marginal likelihood \cite{Rasmussen06}. 
The marginal likelihood in (\ref{eq:marlikelihood}) yields a zero-mean complex proper Gaussian random vector with covariance matrix $\C$, and the log marginal likelihood is
\begin{equationw} \label{eq:CMML}
L=\log p\left(\yv_\n| \X_\n \right)=-\yv_\n\her\C\inv\yv_\n-\log\det\C-{n}\log \pi.
\end{equationw} 
In our model, $\C=\C(\boldsymbol{\theta})$ can be parameterized in terms of a set of hyperparameters, ${\theta}_{i}$, for the kernel $k(\x, \x')$ and the variance of the additive noise, $\sigma_\epsilon^2$, that can be treated as another hyperparameter. 
Note that matrix $\C(\boldsymbol{\theta})$ is a covariance matrix, so it is Hermitian and $L\left(\boldsymbol{\theta}\right)$ is a real function. In the maximization of \EQ{CMML} its derivative is needed. However, the Cauchy-Riemann conditions dictate that nonconstant real-valued functions that are defined on complex domains are not holomorphic, so the complex derivative cannot be used. Instead, we must seek formal or {\it Wirtinger derivatives} \cite{Bouboulis11}, \cite{Hjorungnes07}. Also, since $\C(\boldsymbol{\theta})$ is Hermitian, it has a structure or pattern, therefore generalized complex-valued matrix derivatives need to be considered as follows \cite{Hjorungnes08}. Define 
 \begin{equation} \label{eq:CMML2}
g(\hat\C,\hat\C^{*})=-\yv_\n\her\hat\C\inv\yv_\n-\log\det\hat\C-{n}\log \pi,
\end{equation} 
where $\hat\C$ is a matrix with independent components, i.e., not Hermitian. It is immediate to see that if $\hat\C$ is substituted with $\C$ in $g(\hat\C,\hat\C^{*})$, the marginal likelihood is produced, i.e.,
\begin{equation}
L\left(\boldsymbol{\theta}\right)=\left.g(\hat\C,\hat\C^{*})\right|_{\hat\C=\C(\boldsymbol{\theta})}.
\end{equation} 
Now, since $\hat\C$ is unpatterned, applying the chain rule leads to 
\begin{align} 
\frac{\partial {L}}{\partial \theta_i} &=\left.\frac{\partial {g(\hat\C,\hat\C^{*})}}{\partial \theta_i}\right|_{\hat\C=\C(\boldsymbol{\theta})} \nonumber\\
&=\sum_{k=1}^{n}{\sum_{l=1}^{n}\left(\left.{\frac{\partial {g(\hat\C,\hat\C^{*})}}{\partial (\hat{\C})_{lk}}}\right|_{\hat\C=\C(\boldsymbol{\theta})}\hspace{-0.2cm}{\frac{\partial {({\C})_{lk}}}{\partial \theta_i}} \right)}\nonumber\\
&={\mathrm{Tr}\left(\left(\left.{\frac{\partial {g(\hat\C,\hat\C^{*})}}{\partial \hat\C}}\right|_{\hat\C=\C(\boldsymbol{\theta})}\right)^\top{\frac{\partial {\C}}{\partial \theta_i}}\right)}.\label{eq:derivad}
\end{align}

The derivative of the first term of $g(\hat\C,\hat\C^{*})$ in (\ref{eq:CMML2}) yields
\begin{align} \label{eq:1term}
\frac{\partial {}}{\partial \hat\C}\left(-\yv_\n\her\hat\C\inv\yv_\n\right)&=-\frac{\partial {}}{\partial \hat\C}\mathrm{Tr}\left(-\yv_\n\her\hat\C\inv\yv_\n\right)\nonumber\\
&=(\hat\C^{\top})^{-1}\left(\yv_\n\yv_\n\her\right)^\top(\hat\C^{\top})^{-1}.
\end{align}
The derivative of the second term of of $g(\hat\C,\hat\C^{*})$ in (\ref{eq:CMML2}) yields
\begin{equation} \label{eq:2term}
\frac{\partial {}}{\partial \hat\C}\left(-\log\det\hat\C\right)=-(\hat\C^{\top})^{-1}.
\end{equation}
Substitution of (\ref{eq:1term}) and (\ref{eq:2term}) in (\ref{eq:derivad}) yield
\begin{align} \label{eq:gradientefinal}
\frac{\partial {L}}{\partial \theta_i} ={\mathrm{Tr}\left(\left(\C^{-1}\yv_\n\yv_\n\her\C^{-1}-\C^{-1}\right){\frac{\partial {\C}}{\partial \theta_i}}\right)},
\end{align}
where now the pattern of matrix $\C$ does not yield further simplifications. 
The term ${\partial {\C}}/{\partial \theta_i}$ depends on the chosen covariance function.

\section{experiments}\LABSEC{Exp}

\subsection{Non-null Cross-covariances}

We propose the following example where we randomly sampled a proper complex GP, to later learn it. We generated samples for both the real and the imaginary parts in $[-6,5]$. The GP covariance matrix used was the one in \EQ{kerfil}, constructed from the corresponding filters in \EQ{fil}. The kernel hyperparameters were set as $\gamma=1.125$ and $\mu=2+2\j$. Values $v\rr$ and $v\rrjj$ were set to 1. The real part of the sample function is shown in \FIG{fig1} (top), while the imaginary part is included in \FIG{fig2} (top). Circular complex Gaussian noise with $\sigma_{\epsilon}=0.1$ was added to represent measurement uncertainty and $\n=200$ training noisy samples were randomly chosen. These samples has been depicted as circles in \FIG{fig1} and \FIG{fig2} (top). 
Maximization of the log marginal likelihood in (\ref{eq:CMML}) using (\ref{eq:gradientefinal}) yielded the following estimated values of the hyperparameters: $\hat\gamma=1.0857$, $\hat\mu=1.9671+\j1.8690$, and $\hat\sigma_{\epsilon}=0.0984$. Then, we found the mean (\ref{eq:mediaCPGP}) and variance (\ref{eq:varCPGP}) of the predictive distribution using those training samples and the estimated values of the hyperparameters. The real and the imaginary parts of the predictive mean (\ref{eq:mediaCPGP}) are depicted in \FIG{fig1} (bottom) and \FIG{fig2} (bottom), respectively. The mean squared error of the estimation was $10\log_{10}MSE =-12.52$ dB, computed for 6400 inputs. 

To further study the accuracy of the prediction, we include in \FIG{fig3} and \FIG{fig4} two slices of the surfaces in \FIG{fig1} and \FIG{fig2}, respectively. In \FIG{fig3} the real part of the sample function of the process is plotted (dashed line) versus the real part of the input, the imaginary part of the input was fixed to the value $\Im(x)=4.4430$. The real part of the prediction in (\ref{eq:mediaCPGP}) is depicted in thick red line, along with the grey shaded area that represents the pointwise mean plus and minus two times the standard deviation. The blue circles mark the training samples. Similarly, in \FIG{fig4} the imaginary part of the input was fixed to the value $\Im(x)=-0.5696$ and the imaginary part of the sample function of the process is plotted (dashed line) versus the real part of the input. Again, the imaginary part of the prediction and the training samples are depicted. Four instances of the posterior (\ref{eq:posteriorf}) are plotted in thin blue lines in Figs. \ref{fig:fig3} and \ref{fig:fig4}.
\begin{figure}[tb!]
\begin{center}
\includegraphics[width=8.6cm, draft=false]{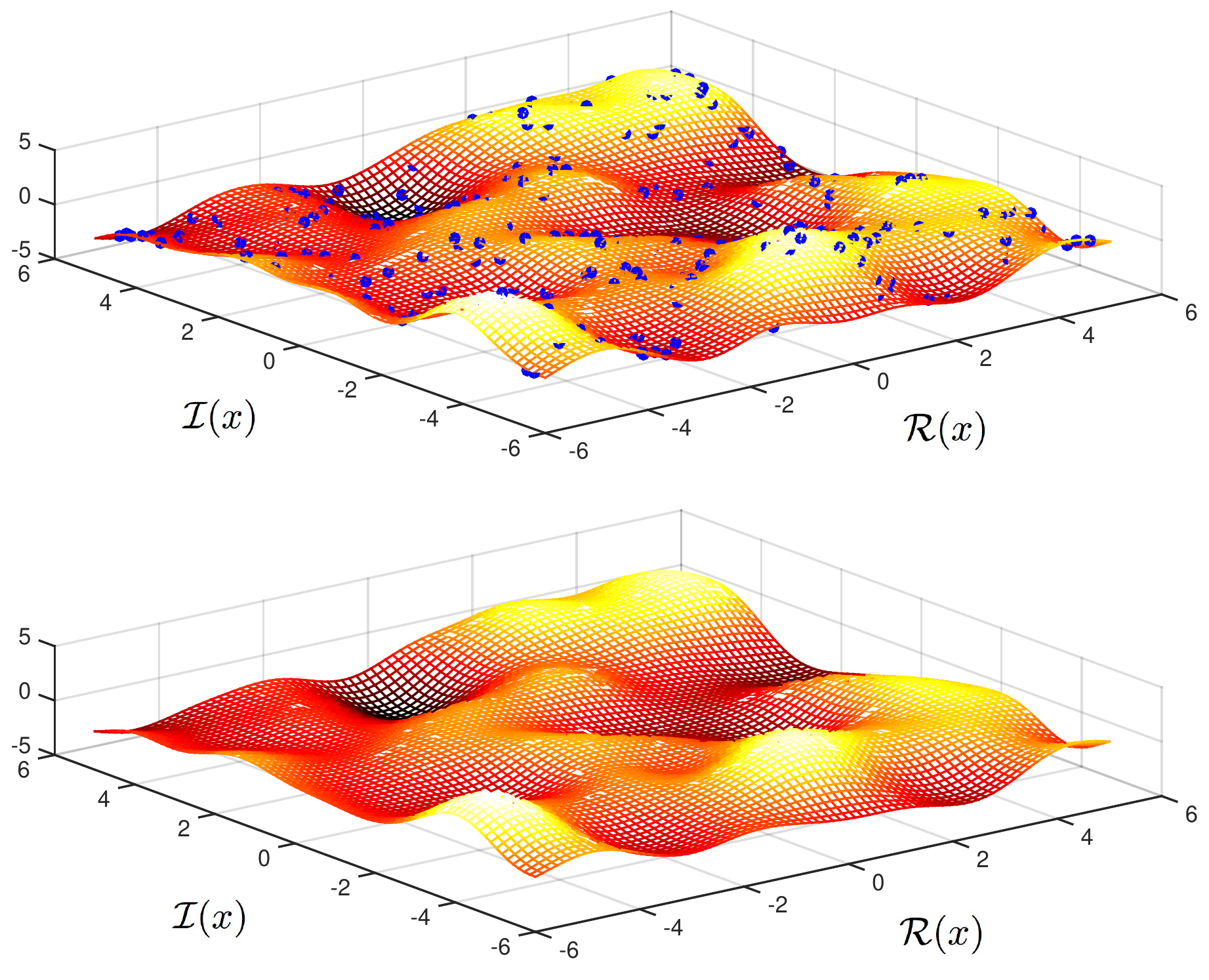}
\end{center}
\vspace*{-.6cm}
\caption{Real part of the output $y(\x)$ (top) and real part of the mean estimation (\ref{eq:mediaCPGP}) (bottom) versus the real and imaginary parts of the input, $\Re(x)$ and $\Im(x)$, respectively. The training samples are depicted as blue circles.}
\LABFIG{fig1}
\end{figure}

\begin{figure}[tb!]
\begin{center}
\includegraphics[width=8.6cm, draft=false]{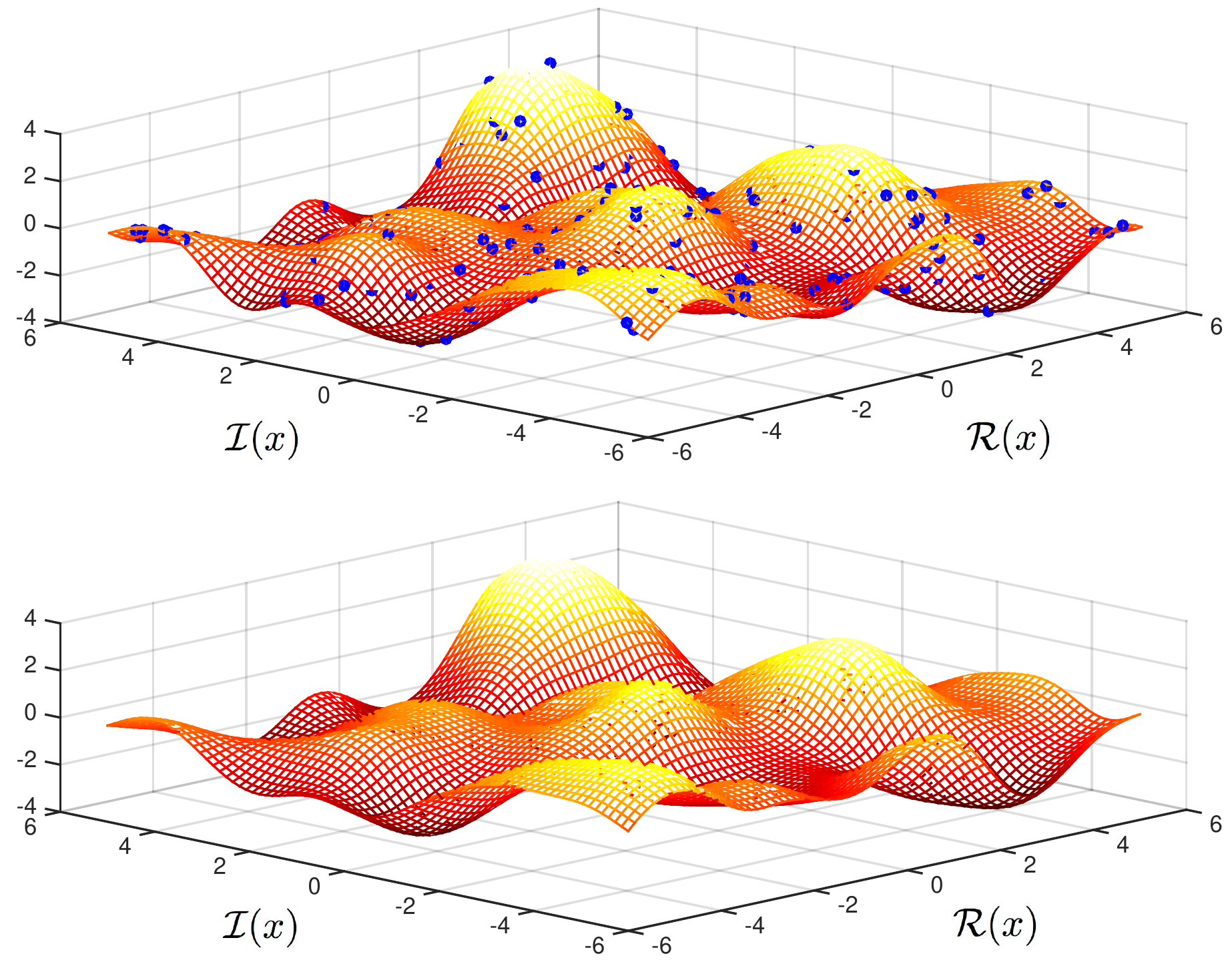}
\end{center}
\vspace*{-.6cm}
\caption{Imaginary part of the output $y(\x)$ (top) and imaginary part of the mean estimation (\ref{eq:mediaCPGP}) (bottom) versus the real and imaginary parts of the input, $\Re(x)$ and $\Im(x)$, respectively. The training samples are depicted as blue circles.}
\LABFIG{fig2}
\end{figure}
\begin{figure}[tb!]
\begin{center}
\includegraphics[width=8.5cm, draft=false]{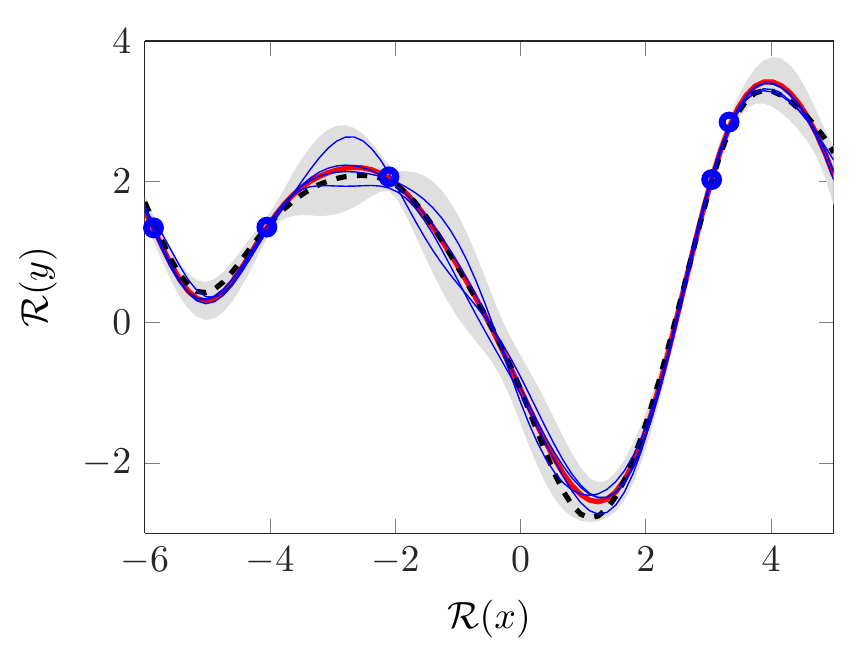}
\end{center}
\vspace*{-.6cm}
\caption{Real part of the output (thick black dashed line) and real part of the predictive mean (\ref{eq:mediaCPGP}) (thick red line) versus $\Re(x)$ for $\Im(x)=4.4430$. Training samples are depicted as blue circles. Four instances of the posterior (\ref{eq:posteriorf}) are plotted in thin blue lines.}
\LABFIG{fig3}
\end{figure}
\begin{figure}[tb!]
\begin{center}
\includegraphics[width=8.5cm, draft=false]{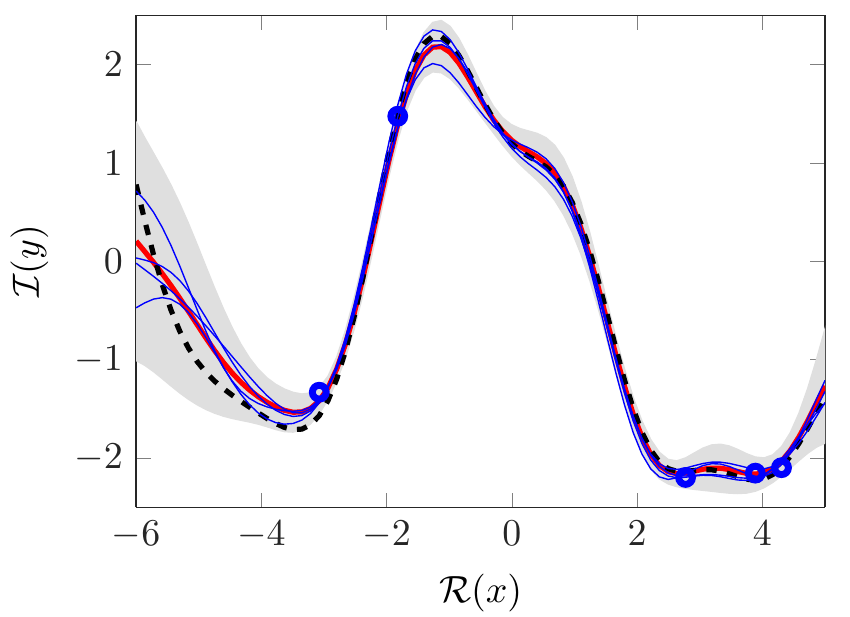}
\end{center}
\vspace*{-.6cm}
\caption{Imaginary part of the output (thick black dashed line) and imaginary part of the predictive mean (\ref{eq:mediaCPGP}) (thick red line) versus $\Re(x)$ for $\Im(x)=-0.5696$. Training samples are depicted as blue circles. Four instances of the posterior (\ref{eq:posteriorf}) are plotted in thin blue lines.}
\LABFIG{fig4}
\end{figure}

%
%

\subsection{Null Cross-covariances}
The performance of the proposed proper complex GPR was tested in the context of a nonlinear channel equalization task. Two nonlinear channels were considered, as in \cite{Bouboulis11} and \cite{Bouboulis12}. Both channels consisted of a linear filter $t(n)=(-0.9+0.8\j)\cdot s(n)+(0.6-0.7\j)\cdot s(n-1)$ and a memoryless nonlinearity. The nonlinearity was $q(n)=t(n)+(0.1+0.15\j)\cdot t^{2}(n)+ (0.06+0.05\j)\cdot t^{3}(n)$ for the first case (labeled as {\it soft nonlinear channel}), and $q(n)=t(n)+(0.2+0.25\j)\cdot t^{2}(n)+ (0.12+0.09\j)\cdot t^{3}(n)$ for the second case (labeled as {\it strong nonlinear channel}). The input signals had the form $s(n)=0.70(\sqrt{1-\rho^{2}}X(n)+\j\rho Y(n))$, as in \cite{Bouboulis11} and \cite{Bouboulis12}, and $X(n)$ and $Y(n)$ were Gaussian random variables. Note that the real and the imaginary parts of the input signals were generated independently and, therefore, had null cross-covariances. Also note that the input signals are circular for $\rho=1/\sqrt{2}$ and highly noncircular if $\rho$ approaches $0$ or $1$. At the receiver end of the channel, the signal $q(n)$ was corrupted by additive white circular Gaussian noise with the SNR set to 16 dB.

The aim of a channel equalization task is to construct an inverse filter, which acts on the received signal $r(t)$ and reproduces the original input signal $s(n)$ as close as possible. To this end, the inputs to the equalizer were the sets of samples $\vect{x}=[r(n+D),r(n+D-1),\cdots,r(n+D-L+1)]^{\top}$, where $L>0$ is the filter length and $D$ is the equalization time delay. 

Experiments were conducted as in \cite{Bouboulis11} and \cite{Bouboulis12}, where $L=5$ and $D=2$, on a set of 3000 samples of the input signal considering both the circular and the noncircular ($\rho=0.1$) cases and the described nonlinear channels ({soft} and {strong}). In all cases the results were averaged over 100 trials where the input signals $s(n)$ and noise output were generated randomly. In Figs. \ref{fig:fig5} to \ref{fig:fig8} we include the MSE along training samples for several methods and the channels and inputs described. As in \cite{Bouboulis11} and \cite{Bouboulis12} the predicted outputs for the training inputs were used to compute the error. The MSE value depicted for each sample is the averaged MSE for the last $100$ samples as in \cite{boubouliscode}\footnote{In the code the average was performed for the last 500 samples.}. 
%
For the sake of comparison we depict the results for the NCLMS, the NCKLMS2 in \cite{Bouboulis11} and the ACKLMS \cite{Bouboulis12} algorithms, that use the complex Gaussian kernel in \EQ{RK}. We used the code available in \cite{boubouliscode} to run these algorithms. For the ACKLMS algorithm all the parameters were set to the values described in \cite{Bouboulis12} ($\gamma=10^{2}$ and the step update parameter $1/8$). Also, for the NCKLMS2 algorithm and the soft nonlinear channel case the parameters were set to the values described in \cite{Bouboulis12} ($\gamma=10^{2}$ and the step update parameter $1/8$), whereas for the strong nonlinear channel case the parameters were set to the values described in \cite{Bouboulis11} ($\gamma=5^{2}$ and the step update parameter $1/4$). The novelty criterion was used for the sparsification of both the NCKLMS2 and the ACKLMS algorithms with $\delta_{1}=0.15$ and $\delta_{2}=0.2$. 
 In Figs. \ref{fig:fig7} and \ref{fig:fig8} we observe stability problems in the learning process of both NCKLMS2 and ACKLMS, due to the kernel used. This problem was alleviated by using sparsification. We also used the independent kernel (\ref{eq:indepCGK}) with $\kappa_{\Rext}$ being the real Gaussian kernel in the NCKLMS2 approach, as proposed in \cite{Tobar12}. We labeled this algorithm as NCKLMS2-i. The tunable parameter was set to $\gamma=5^{2}$ and the step update parameter to $1/8$. The novelty criterion was again used for the sparsification. 

In this paper we propose the mean (\ref{eq:mediaCPGP}) of the predictive distribution for the proper complex GPR as the equalizer output. We used the kernel in \EQ{complexCV} where $k\rrrr(\x,\x')$$=k\jjjj(\x,\x')$$=\k_{G}(\x,\x')/2$ with $\k_{G}(\x,\x')$ as in (\ref{eq:expkernel}), and $k\rrjj(\x,\x')=0$, since the real and the imaginary parts of $s(n)$ had null cross-covariances. Two approaches were simulated. In one procedure, which we labeled as opt-CGPR, $1000$ input-output samples were randomly chosen to tune the kernel hyperparameter $\gamma$ so that the best possible results were obtained. The noise variance $\sigma_{\epsilon}^{2}$ in the kernel was set to the valued used in the experiments. Then, with those hyperparameter values, for each new input $\vect{x}$ the equalizer predicted the corresponding $s(n)$ as in (\ref{eq:mediaCPGP}) using all the previous input-output pairs as training set. It can be observed through the figures the remarkable good results of this proposal in all the cases, soft or strong nonlinear channels and circular or noncircular signals. Also, note that the MSE increases with the number of samples until reaches steady estate, as CGPR generalizes for an increasing number of input-output pairs.
In a second approach, labeled as CGPR, we tried to check the hyperparameter estimation capabilities of the proper complex GPR by means of the maximization of the log marginal likelihood (\ref{eq:CMML}) using (\ref{eq:gradientefinal}). The first $250$ samples were used as training set to estimate the hyperparameters ($\gamma$ and $\sigma_{\epsilon}$). Then, with those values for $\gamma$ and $\sigma_{\epsilon}^{2}$ fixed, and for each new input $\vect{x}$ the equalizer predicted the corresponding $s(n)$ as in (\ref{eq:mediaCPGP}) using all the previous input-output pairs as training set. The learning curves for this procedure are depicted in Figs. \ref{fig:fig5} to \ref{fig:fig8}. The results for this procedure, CGPR, not far from that of the opt-CGPR, considerably outperform the NCKLMS2 and ACKLMS and illustrate the good estimation of the hyperparameters. 

Finally, we proposed to improve the NCKLMS2 by using the kernels derived in \SEC{Cov}, in particular the kernel in (\ref{eq:expkernel}) with null imaginary part. This algorithm is labeled as NCKLMS2-G. We set all the parameters for this algorithm to the same values indicated in \cite{Bouboulis11} ($\gamma=5^{2}$ and the step update parameter $1/4$), and the novelty criterion was again used for the sparsification. While the NCKLMS2-i algorithm exhibits improved performance compared to the NCKLMS2 or ACKLMS algorithms, the NCKLMS2-G algorithm outperforms all of them with a smooth learning curve under all conditions. This proves the proposed complex kernel design procedure to be a valuable one. In this equalization problem the cross-covariance between real and imaginary parts of the signals to-be-learned, $s(n)$, is null. Therefore, we set the imaginary part of the kernel to zero with much better results. Also, the measure of similarity of inputs through the simple norm of the complex difference between inputs endows the kernel with useful properties such as isotropy and stationarity, better fitting the underlying model. 
\begin{figure}[tb!]
\begin{center}
\includegraphics[width=8.6cm, draft=false]{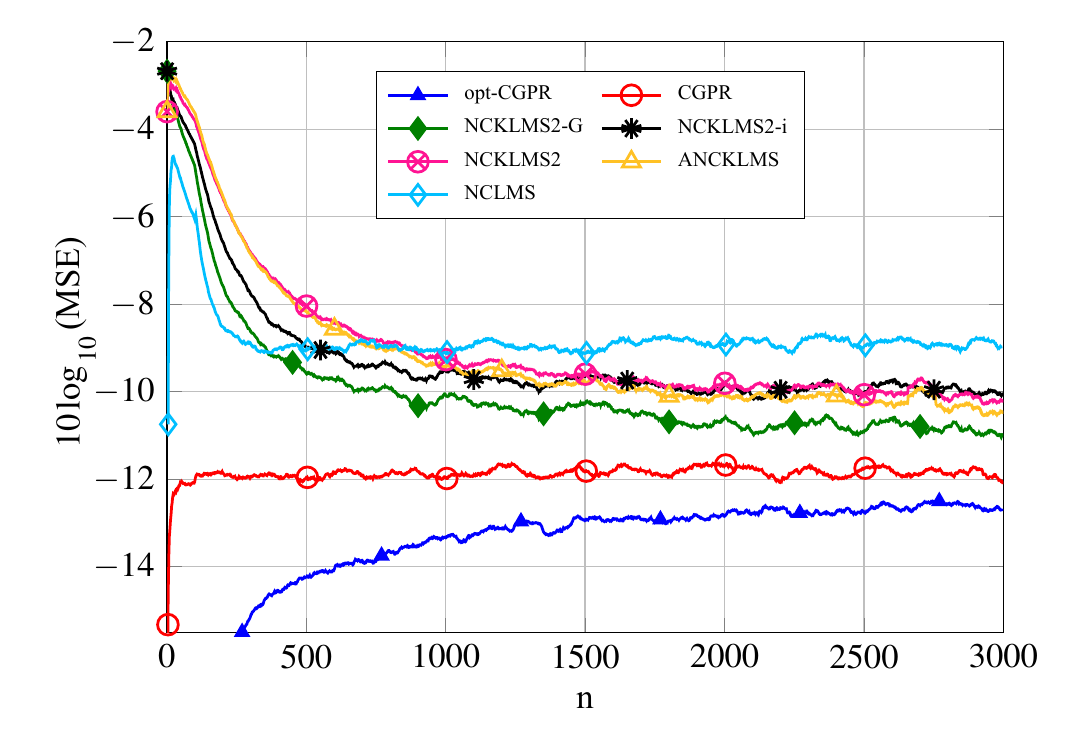}
\end{center}
\caption{Learning curves for NKCLMS2, NCKLMS2-i, NKCLMS2-G, AKCLMS, the proper complex GPR (CGPR) and the optimized CGPR (opt-CGPR) for the soft nonlinear channel equalization problem for the circular input case.}
\LABFIG{fig5}
\end{figure}
\begin{figure}[tb!]
\begin{center}
\includegraphics[width=8.6cm, draft=false]{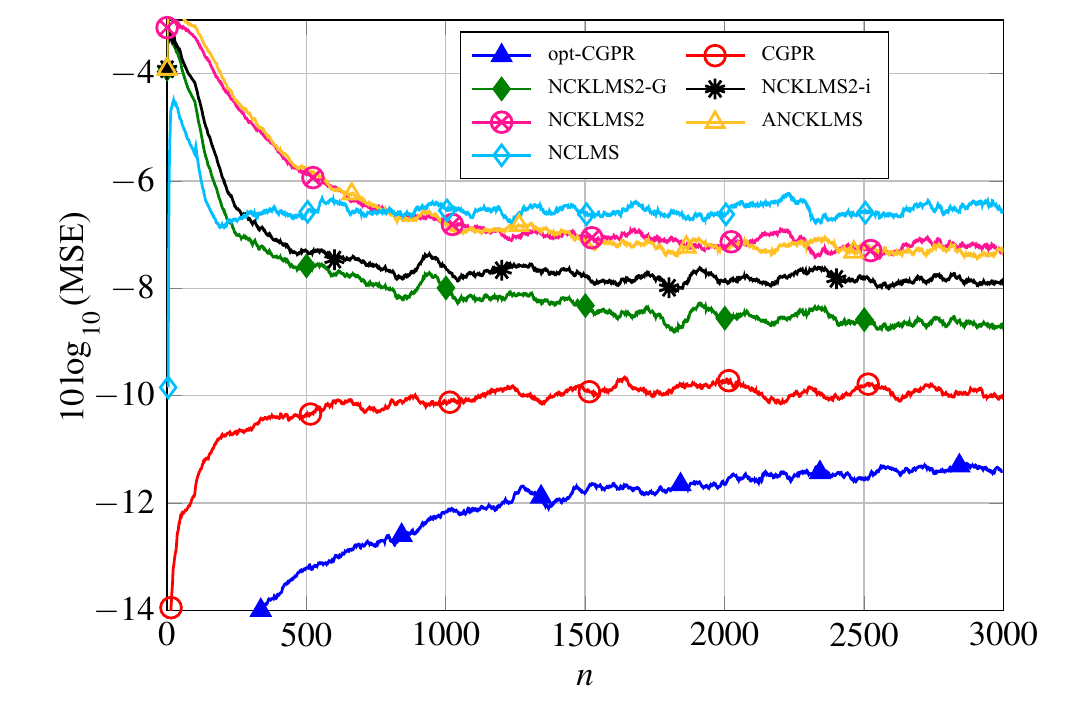}
\end{center}
\caption{Learning curves for NKCLMS2, NCKLMS2-i, NKCLMS2-G, AKCLMS, the proper complex GPR (CGPR) and the optimized CGPR (opt-CGPR) for the strong nonlinear channel equalization problem for the circular input case.}
\LABFIG{fig6}
\end{figure}

\begin{figure}[tb!]
\begin{center}
\includegraphics[width=8.6cm, draft=false]{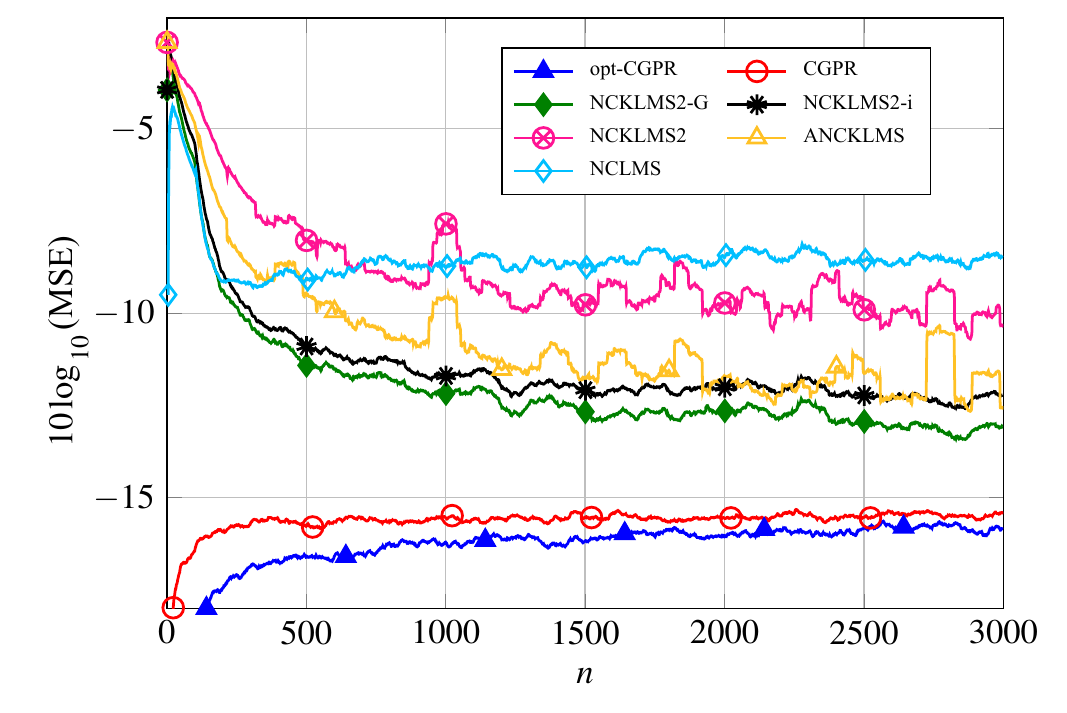}
\end{center}
\vspace*{-.6cm}
\caption{Learning curves for NKCLMS2, NCKLMS2-i, NKCLMS2-G, AKCLMS, the proper complex GPR (CGPR) and the optimized CGPR (opt-CGPR) for the soft nonlinear channel equalization problem for the noncircular input case ($\rho=0.1$).}
\LABFIG{fig7}
\end{figure}
\begin{figure}[tb!]
\begin{center}
\includegraphics[width=8.6cm, draft=false]{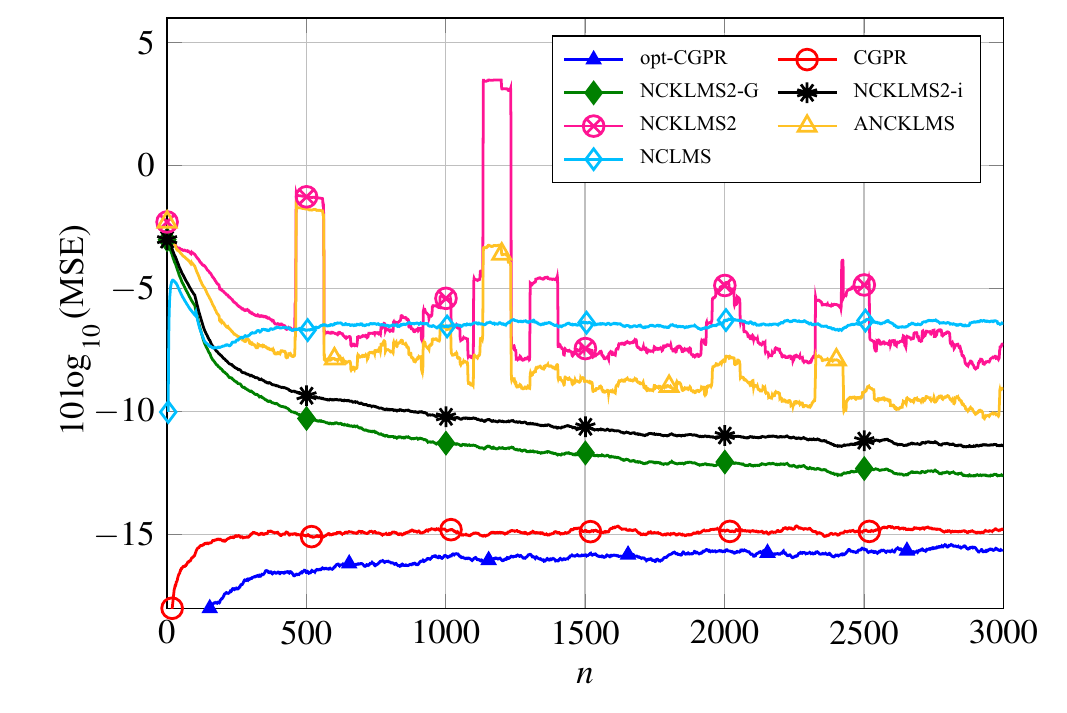}
\end{center}
\vspace*{-.6cm}
\caption{Learning curves for NKCLMS2, NCKLMS2-i, NKCLMS2-G, AKCLMS, the proper complex GPR (CGPR) and the optimized CGPR (opt-CGPR) for the strong nonlinear channel equalization problem for the noncircular input case ($\rho=0.1$).}
\LABFIG{fig8}
\end{figure}

\section{Conclusions}

In this paper we prove that the kernels for complex-valued regression proposed in the literature correspond to \emph{proper} complex outputs. Since proper complex-valued Gaussian processes have been widely studied, we propose to resort to the Gaussian processes for regression to study this problem. On the one hand, we end with a new solution that it is endowed with the advantages of GPs: a probabilistic output and the availability of estimating the hyperparameters by optimization. On the other, we derive an expression for the reproducing kernel in the proper complex case. We conclude that it is complex valued if the cross-covariance of the real and imaginary parts of the outputs is not null. The imaginary part of the covariance matrix is skew-symmetric, and the full matrix must be semidefinite positive, but the kernel used in the real and imaginary parts need not to be the same. Besides, if the cross-covariance cancels, the kernel is real-valued. The MOL point of view bears this out. Therefore, we prove the kernel to exhibit a much more flexible structure than proposed in previous works. We end designing a kernel where the use of an appropriate metric yields an isotropic and stationary kernel. The proposed methods developed by fully using these key results exhibit a remarkable good performance compared to previous ones. 


\appendix[Covariance Function]
We follow here a procedure similar to that in \cite{Boyle05}. Consider two independent, real, stationary, Gaussian white noise processes $S\rr(\x)$ and $S\jj(\x)$, where $\x\in\CN^\d$, producing an output $Y(\x)=U(\x)+W(\x)$, where $W(\x)$ is a stationary Gaussian white noise, and $U(\x)$ is defined by the sum of convolutions
\begin{align}
U(\x)&=\left(h\rr(\x)+\j h\rrjj(\x)\right) \star S\rr(\x)+\left(h\rr(\x)+\j h\jjrr(\x)\right) \star S\jj(\x)\nonumber\\
&=\sum_{m=1}^4 h_m(\x) \star S_m(\x),
\end{align}
where we have used the following notation: $h_1(\x)=h\rr(\x)$, $h_2(\x)=\j h\rrjj(\x)$, $h_3(\x)=h\rr(\x)$, $h_4(\x)=\j h\jjrr(\x)$, $S_1(\x)=S_2(\x)= S\rr(\x)$, and $S_3(\x)=S_4(\x)= S\jj(\x)$.
The covariance of $Y(\x)$ is derived as follows:
\begin{align}
\C(\x_a,\x_b)=\C_{U}(\x_a,\x_b)+\sigma_W^2\delta_{ab},
\end{align}
where $\sigma_W^2$ is the variance of $W(\x)$, and
\begin{align} \LABEQ{app2}
\C_{U}(\x_a,\x_b)&=\E\left[U\rr(\x_a)U\rr^{*}(\x_b)\right]\nonumber\\
&=\E\left[\sum_{m=1}^4\int_{\CN^\d}{h_m(\mathbf{\alpha})S_m(\x_a-\mathbf{\alpha})\dfd{\d}{\mathbf{\alpha}}}\right.\nonumber\\&\cdot\left.\sum_{n=1}^4\int_{\CN^\d}{h_n^{*}(\mathbf{\beta})S_n(\x_b-\mathbf{\beta})\dfd{\d}{\mathbf{\beta}}}\right]
\nonumber\\
&=\sum_{m=1}^4\sum_{n=1}^4\left\{\int_{\CN^\d}\int_{\CN^\d}h_m(\mathbf{\alpha})h_n^{*}(\mathbf{\beta})\right.\nonumber\\&\cdot\left.\E\left[S_m(\x_a-\mathbf{\alpha})S_n(\x_b-\mathbf{\beta})\right]\dfd{\d}{\mathbf{\alpha}}\dfd{\d}{\mathbf{\beta}}\right\}.
\end{align}
Since $S_1(\x)=S_2(\x)= S\rr(\x)$, and $S_3(\x)=S_4(\x)= S\jj(\x)$, processes $S_m(\x_a-\mathbf{\alpha})$ and $S_n(\x_b-\mathbf{\beta})$ covary only if $m,n\in\{1,2\}$ or $m,n\in\{3,4\}$, and $(\x_a-\mathbf{\alpha})=(\x_b-\mathbf{\beta})$. In such cases, $\E\left[S_m(\x_a-\mathbf{\alpha})S_n(\x_b-\mathbf{\beta})\right]=\delta(\mathbf{\alpha}-(\x_a-\x_b+\mathbf{\beta}))$, and the integrals in \EQ{app2} yield
\begin{align}
&\int_{\CN^\d}\int_{\CN^\d}h_m(\mathbf{\alpha})h_n^{*}(\mathbf{\beta})\delta(\mathbf{\alpha}-(\x_a-\x_b+\mathbf{\beta}))\dfd{\d}{\mathbf{\alpha}}\dfd{\d}{\mathbf{\beta}}\nonumber\\
&=\int_{\CN^\d}h_m(\mathbf{\beta}+(\x_a-\x_b))h_n^{*}(\mathbf{\beta})\dfd{\d}{\mathbf{\beta}}\nonumber\\
&=\int_{\CN^\d}h_m(\mathbf{\beta}+\mathbf{d}_\x)h_n^{*}(\mathbf{\beta})\dfd{\d}{\mathbf{\beta}},
\end{align}
where $\mathbf{d}_\x=\x_a-\x_b$.
Hence,
\begin{align} \LABEQ{app1}
\C_{U}(\x_a,\x_b)&=\sum_{m=1}^2\sum_{n=1}^2\int_{\CN^\d}h_m(\mathbf{\beta}+\mathbf{d}_\x)h_n^{*}(\mathbf{\beta})\dfd{\d}{\mathbf{\beta}}\nonumber\\
&+\sum_{m=3}^4\sum_{n=3}^4\int_{\CN^\d}h_m(\mathbf{\beta}+\mathbf{d}_\x)h_n^{*}(\mathbf{\beta})\dfd{\d}{\mathbf{\beta}}.
\end{align}
If we set the kernels to parameterized exponentials as $h_i(\x)=(\j)^{c_{i}}v_{i}\exp(-(\x-\mu_i)\her(\x-\mu_i)/\gamma)$, integrals in \EQ{app1} are as follows
 \begin{align}\LABEQ{app3}
&\int_{\CN^\d}h_m(\mathbf{\beta}+\mathbf{d}_\x)h_n^{*}(\mathbf{\beta})\df{\mathbf{\beta}^\d}\nonumber
\\=&(\j)^{c_{m}}(-\j)^{c_{n}}v_{m}v_{n}\int_{\CN^\d}\exp\left(-\frac{(\mathbf{\beta}-\mu_n)\her(\mathbf{\beta}-\mu_n)}{\gamma}\right)\nonumber\\\cdot&\exp\left(-\frac{(\mathbf{\beta}+\mathbf{d}_\x-\mu_m)\her(\mathbf{\beta}+\mathbf{d}_\x-\mu_m)}{\gamma}\right)\df{\mathbf{\beta}^\d}\nonumber\\
=&(\j)^{c_{m}}(-\j)^{c_{n}}v_{m}v_{n}\left(\int_{\CN^\d}\exp\left(-\frac{(\mathbf{\beta}-\hat\beta)\her(\mathbf{\beta}-\hat\beta)}{0.5\gamma}\right)\df{\mathbf{\beta}^\d}\right)\nonumber\\&\cdot\exp\left(-\frac{(\mathbf{d}_\x-\mu_m+\mu_n)\her(\mathbf{d}_\x-\mu_m+\mu_n)}{2\gamma}\right)\nonumber\\
=&(\j)^{c_{m}}(-\j)^{c_{n}}\left(\frac{\pi\gamma}{2}\right)^{\d}v_{m}v_{n}\nonumber\\\cdot&\exp\left(-\frac{(\mathbf{d}_\x-\mu_m+\mu_n)\her(\mathbf{d}_\x-\mu_m+\mu_n)}{2\gamma}\right),
\end{align}
where $\hat\beta=0.5(\mu_n-(\mathbf{d}_\x-\mu_m))$. 

We propose the following parameter values. For $h_1(\x)=h_3(\x)=h\rr(\x)$, we set $v_{1}=v_{3}=v\rr$, $\mu_{1}=\mu_{3}=0$ and $c_{1}=c_{3}=0$. For $h_2(\x)=\j h\rrjj(\x)$, we set $\mu_{2}=\mu$, $v_{2}=v\rrjj$ and $c_{2}=1$. And for $h_4(\x)=\j h\jjrr(\x)$, we set $\mu_{4}=-\mu$, $v_{4}=-v\rrjj$ and $c_{4}=1$. By making use of this values and \EQ{app3}, \EQ{app1} yields
\begin{align} 
&\C_{U}(\x_a,\x_b)=\C_{U}(\mathbf{d}_\x)\nonumber\\
&=\left(\frac{\pi\gamma}{2}\right)^{\d}v\rr^{2}\exp\left(-\frac{\mathbf{d}_\x\her\mathbf{d}_\x}{2\gamma}\right)\nonumber\\
&-\j\left(\frac{\pi\gamma}{2}\right)^{\d}v\rr v\rrjj\exp\left(-\frac{(\mathbf{d}_\x+\mu)\her(\mathbf{d}_\x+\mu)}{2\gamma}\right)\nonumber\\
&+\j\left(\frac{\pi\gamma}{2}\right)^{\d}v\rr v\rrjj\exp\left(-\frac{(\mathbf{d}_\x-\mu)\her(\mathbf{d}_\x-\mu)}{2\gamma}\right)\nonumber\\
&+\left(\frac{\pi\gamma}{2}\right)^{\d}v\rrjj^{2}\exp\left(-\frac{\mathbf{d}_\x\her\mathbf{d}_\x}{2\gamma}\right)+\left(\frac{\pi\gamma}{2}\right)^{\d}v\rr^{2}\exp\left(-\frac{\mathbf{d}_\x\her\mathbf{d}_\x}{2\gamma}\right)\nonumber\\
&+\j\left(\frac{\pi\gamma}{2}\right)^{\d}v\rr v\rrjj\exp\left(-\frac{(\mathbf{d}_\x-\mu)\her(\mathbf{d}_\x-\mu)}{2\gamma}\right)\nonumber\\
&-\j\left(\frac{\pi\gamma}{2}\right)^{\d}v\rr v\rrjj\exp\left(-\frac{(\mathbf{d}_\x+\mu)\her(\mathbf{d}_\x+\mu)}{2\gamma}\right)\nonumber\\
&+\left(\frac{\pi\gamma}{2}\right)^{\d}v\rrjj^{2}\exp\left(-\frac{\mathbf{d}_\x\her\mathbf{d}_\x}{2\gamma}\right)\nonumber\\
&=\left(\frac{\pi\gamma}{2}\right)^{\d}(2v\rr^{2}+2v\rrjj^{2})\exp\left(-\frac{\mathbf{d}_\x\her\mathbf{d}_\x}{2\gamma}\right)\nonumber\\
&+\j\left(\frac{\pi\gamma}{2}\right)^{\d}2v\rr v\rrjj\left(\exp\left(-\frac{(\mathbf{d}_\x-\mu)\her(\mathbf{d}_\x-\mu)}{2\gamma}\right)\right.\nonumber\\
&-\left.\exp\left(-\frac{(\mathbf{d}_\x+\mu)\her(\mathbf{d}_\x+\mu)}{2\gamma}\right)\right).
\end{align}


%


\ifCLASSOPTIONcaptionsoff
  \newpage
\fi



%

\bibliographystyle{IEEEtran}
\bibliography{IEEEabrv,CGPR,murilloGP,SSCDMA}

%
%

%








\end{document}